%% file: main.tex
\documentclass[9pt, conference]{IEEEtran}
\IEEEoverridecommandlockouts
\usepackage{cite}
\usepackage{amsmath,amssymb,amsfonts}
\usepackage{algorithmic}
\usepackage{graphicx}
\usepackage{textcomp}
\usepackage{xcolor}
\def\BibTeX{{\rm B\kern-.05em{\sc i\kern-.025em b}\kern-.08em
    T\kern-.1667em\lower.7ex\hbox{E}\kern-.125emX}}

\usepackage{tikz}
\usepackage{calc}

\usepackage[noabbrev]{cleveref}

\usepackage{subfigure}
\usepackage{booktabs} 
\usepackage{caption}
\usepackage{stfloats}
\usepackage{wrapfig}

\usepackage{pifont}
\newcommand\mynuma[1]{\ifcase#1 \or \ding{172}\or \ding{173}\or
  \ding{174}\or \ding{175}\or \ding{176}\or \ding{177}%
  \or \ding{178}\or \ding{179}\or \ding{180}\or \ding{181}\else *\fi\relax}

\newcommand\mynumb[1]{\ifcase#1 \or \ding{182}\or \ding{183}\or
  \ding{184}\or \ding{185}\or \ding{186}\or \ding{187}%
  \or \ding{188}\or \ding{189}\or \ding{190}\or \ding{191}\else *\fi\relax}

\usepackage{nccmath}  
\usepackage{bbm}
\usepackage{amsmath}
\usepackage{amssymb}
\usepackage{mathtools}
\usepackage{amsthm}
\usepackage{multirow}
\usepackage{booktabs}
\usepackage{xcolor}
\usepackage{booktabs}
\usepackage{enumitem}
\usepackage{array}
\usepackage{wrapfig}
\usepackage{fdsymbol}

\definecolor{Note_color}{rgb}{0.0, 0.0, 1.0}

\DeclareMathOperator*{\argmin}{arg\,min}

\begin{document}

\title{Robust Tickets Can Transfer Better: Drawing More Transferable Subnetworks in Transfer Learning}

\author{
\IEEEauthorblockN{Yonggan Fu$^1$, Ye Yuan$^1$, Shang Wu$^2$, Jiayi Yuan$^2$, Yingyan (Celine) Lin$^1$}
\IEEEauthorblockA{$^1$\textit{Georgia Institute of Technology}, $^2$\textit{Rice University}}
\IEEEauthorblockA{\textit{\{yfu314, eiclab.gatech, celine.lin\}@gatech.edu \, \{sw99, jy101\}@rice.edu}}
\vspace{-1.5em}
}

\maketitle

\input{Sections/0-Abstract}

\input{Sections/1-Introduction}

\input{Sections/2-Method}

\input{Sections/3-Experiment}
\input{Sections/4-Related-Work}

\input{Sections/5-Conclusion}

\bibliographystyle{IEEEtranS}
\bibliography{ref}

\end{document}

%% file: Sections/0-Abstract.tex
\begin{abstract}

Transfer learning leverages feature representations of deep neural networks (DNNs) pretrained on source tasks with rich data to empower effective finetuning on downstream tasks. However, the pretrained models are often prohibitively large for delivering generalizable representations, which limits their deployment on edge devices with constrained resources. To close this gap, we propose a new transfer learning pipeline, which leverages our finding that robust tickets can transfer better, i.e., subnetworks drawn with properly induced adversarial robustness can win better transferability over vanilla lottery ticket subnetworks. Extensive experiments and ablation studies validate that our proposed transfer learning pipeline can achieve enhanced accuracy-sparsity trade-offs across both diverse downstream tasks and sparsity patterns, further enriching the lottery ticket hypothesis.

\end{abstract}

%% file: Sections/1-Introduction.tex
\section{Introduction}
\label{sec:intro}

Deep neural networks (DNNs) trained on large-scale datasets have prevailed as state-of-the-art (SOTA) solutions for various cognition tasks across many application domains, thanks to their record-breaking performance. However, the availability of a large amount of training data, one of the major driving forces behind the amazing success of DNNs, is not often possible due to the required high cost of data collection and laborious annotations. Fortunately, recent advances in transfer learning, which leverage rich feature representations learned on source tasks for which big training datasets are available to empower the learning on downstream tasks with limited data resources, have provided a promising data-efficient solution for enhancing the achievable downstream accuracy. For example, a typical solution of transfer learning for boosting task  accuracy on small-scale vision tasks is to finetune the models pretrained on large-scale datasets in a supervised or unsupervised manner.

In parallel, it is increasingly demanded to advance DNN-powered edge applications by leveraging the recent success of transfer learning, i.e., adapting the pretrained models that serve as general feature extractors to various downstream tasks on the edge where collecting high-quality annotated data at scale is difficult or not practical.
Nevertheless, the pretrained models are often prohibitively large in order to ensure generalizable feature representations, which stands at odds with the limited resources available on edge devices like mobile phones. Therefore, it is highly desired to trim down the complexity of large pretrained models while at the same time maintaining their transferability to various downstream tasks, which can be drastically different from the goal of standard model compression which aims to preserve the task accuracy on the same dataset after compression.

To close the aforementioned gap, recent pioneering works~\cite{chen2021lottery,iofinova2021well} have extended the lottery ticket hypothesis~\cite{frankle2018lottery} to pretrained models under a transfer learning setting, i.e., they have shown that there exist subnetworks, which inherit the pretrained model weights as initialization, can match the task accuracy of their dense network counterparts after finetuning on downstream tasks, which shed light on potential opportunities of inducing sparsity into large pretrained models for enhanced efficiency without degrading their transferability. However, one missing piece is that all these previous works directly reuse the metrics for identifying lottery tickets under a standard training setting, e.g., weight magnitudes~\cite{chen2021lottery, iofinova2021well}, which were originally designed for maintaining the accuracy on the same task and thus do not necessarily help preserve the transferability to downstream tasks. We emphasize that given the general existence of lottery tickets in pretrained models, extra priors are required for identifying more transferable ones among them in addition to the aforementioned metrics.

To this end, we ask an intriguing question: ``what kind of priors should we consider for drawing more transferable tickets from pretrained models?" Inspired by recent works~\cite{salman2020adversarially,deng2021adversarial} showing that enhancing adversarial robustness~\cite{goodfellow2014explaining} of pretrained models results in better representations that align well with human perceptions~\cite{deng2021adversarial}, we hypothesize that subnetworks hidden in pretrained models with properly induced adversarial robustness, dubbed \textit{robust tickets}, can win better transferability over vanilla lottery tickets drawn without considering a robustness objective, dubbed \textit{natural tickets}. In other words, the key insight of this work is that adversarial robustness can serve as a proper prior for drawing more transferable tickets from pretrained models for transfer learning. In particular, we summarize our contributions as follows:

\begin{itemize}
    \item We discover that robust tickets can transfer better,
    i.e., properly induced adversarial robustness can serve as a good prior for drawing more transferable subnetworks from pretrained models under a transfer learning setting.  
    
    \item We extensively study different schemes for drawing robust tickets from pretrained models, leading to new pipelines for more effectively transferring decent subnetworks to downstream tasks, which can push forward the achievable transferability-sparsity trade-offs over natural ones.
    
    \item We conduct extensive experiments to understand the properties of robust tickets, and benchmark their effectiveness over natural tickets across different \underline{(1)} datasets and tasks, \underline{(2)} sparsity patterns, \underline{(3)} pretraining schemes, and \underline{(4)} performance metrics, including downstream accuracy, the robustness under adversarial perturbations, and out-of-distribution detection performance.

    \item We empirically analyze the underlying reasons behind the transferability of robust tickets, which is found to be highly correlated to their capability of handling domain gaps, and explore the boundary regarding \underline{(1)} whether and \underline{(2)} when robust tickets could transfer better than natural tickets.

\end{itemize}

We believe this work has not only provided a new perspective that complements the lottery ticket hypothesis for transfer learning, but also opened up a new angle for empowering transfer learning on edge devices toward enhanced accuracy-efficiency trade-offs.

%% file: Sections/2-Method.tex
\section{Hypothesis and Methodology}
\label{sec:method}

\subsection{Key Hypothesis}
We hypothesize that subnetworks drawn from pretrained models with properly induced adversarial robustness (i.e., robust tickets), can win better transferability in terms of the achievable accuracy on downstream tasks, as compared to vanilla lottery tickets drawn without considering any robustness objective (natural tickets). 
This is inspired by recent observations~\cite{salman2020adversarially,deng2021adversarial} showing that adversarial-robustness-aware training enables learning better feature representations, which align well with human perceptions, and thus can enhance the achievable task performance of transfer learning  thanks to the resulting bias of focusing on less difficult inputs with large signal-to-noise ratios and the removal of redundant features~\cite{deng2021adversarial}.
Built upon the aforementioned hypothesis, our work leverages adversarial robustness as a prior for drawing more transferable tickets from pretrained models, considering the general existence of lottery tickets.

\subsection{Drawing Robust Tickets}
\textbf{Overview.} To practically leverage our hypothesis above for discovering highly transferable tickets, it is critical to identify how to induce adversarial robustness as priors during the process of drawing lottery tickets. We find that enhancing the adversarial robustness of the pretrained dense model can serve as a simple but effective way to introduce robustness priors for drawing tickets with better transferability on downstream tasks. This is inspired by  recent observations showing that \underline{(1)} lottery tickets drawn from adversarially trained DNNs can preserve decent adversarial robustness~\cite{li2020towards}, and \underline{(2)} the adversarial robustness of pretrained models on a source task can be inherited to downstream tasks~\cite{yamada2022does}. 

Therefore, we consider a two-stage process for delivering robust tickets: \underline{(1)} inducing adversarial robustness to the dense models when pretraining them on the source tasks, and \underline{(2)} drawing robust tickets from the robustly pretrained dense models resulting from the previous stage, which are then transferred to downstream tasks.

\textbf{Inducing adversarial robustness to pretrained models.} In this work, we mainly apply adversarial training~\cite{madry2017towards}, considering that it is one of the most effective robustifying methods when pretraining a dense model on a source task. Additionally, we also adopt random smoothing~\cite{cohen2019certified} to validate the generality of our discovered insight.

\textbf{Different schemes for discovering robust tickets.}
We adopt different pruning methods to draw robust tickets for fairly benchmarking with their corresponding natural ticket counterparts. In particular, a robust ticket can be formulated as $f(\cdot;  m \odot \theta_{pre})$, where $f$ is the pretrained model parameterized by $\theta_{pre}$ and $m$ is a binary mask for indexing the sparse subnetworks within the pretrained model. We derive $m$ with three different schemes as follows and evaluate the transferability of our proposed robust tickets and commonly adopted natural tickets under these schemes in the experiment section.

\color{black}{\ding{172}} \textbf{One-shot magnitude pruning (OMP):} In OMP, we directly prune weights with the smallest magnitudes based on $||\theta_{pre}||$ toward the target pruning ratio, and identify the model with the remaining weights as the robust ticket, which is further transferred to downstream tasks. 
Note that robust tickets and natural tickets differ in the pretrained weights $||\theta_{pre}||$ during OMP: The former is drawn from robustly pretrained dense models while the latter is drawn from naturally pretrained dense models.

\color{black}{\ding{173}} \textbf{Adversarial iterative magnitude pruning (A-IMP):} Motivated by the success of IMP in discovering lottery tickets across application domains~\cite{frankle2018lottery,chen2021lottery}, we propose an adversarial variant of IMP, dubbed as A-IMP, for drawing robust tickets. Following the common practice of IMP~\cite{frankle2018lottery,chen2021lottery}, after each pruning iteration, weights with the smallest magnitudes as determined by the target pruning ratio of the current iteration will be pruned and then the sparsity of $m$ is further increased. This process is iteratively repeated until reaching the target sparsity. To induce the target robustness prior during iterative pruning of the pretrained model, we modify the training objective to an adversarial formulation that performs a minimax optimization:

\begin{equation} \label{eq:imp_objective}
    \argmin_{\theta} \,\,  \max_{\|\delta\|_{\infty} \leq \epsilon} \ell(f(m \odot \theta, x+\delta), y)
\end{equation} 

\noindent where $x$ and $y$ are the input and label pairs of the target task, $l$ is the corresponding loss function, and $\delta$ is the adversarial perturbation under a norm constraint of $\epsilon$.

\color{black}{\ding{174}} \textbf{Learnable mask pruning (LMP):} LMP directly learns a task-specific mask $m_t$ for each downstream task on top of the pretrained model without further tuning the model weights~\cite{fu2021drawing,ramanujan2020s,fu2022losses}. 
In particular, LMP can be formulated as:

\begin{equation} \label{eq:lmp_objective}
    \argmin_{m_t} \,\,  \ell_t(f(m_t \odot \theta_{pre}, x_t), y_t) \,\,\,\,\, s.t. \,\,\, ||m_t||_0  \leqslant k_t
\end{equation} 

\noindent where $x_t$ and $y_t$ are the input and label pairs of the downstream task $t$ and an $L_0$ constraint is exerted on $m_t$ to ensure the number of its non-zero elements is no more than $k_t$. To optimize $m_t$ in a differentiable manner, following~\cite{ramanujan2020s}, during forward we binarize $m_t$ to $\hat{m}_t$, which approximates the top $k_t$ elements of $m_t$ using 1 and otherwise 0, while during backward all the elements of $m_t$ are updated via straight-through estimation, i.e., $\frac{\partial l_t}{\partial m_t}\approx\frac{\partial l_t}{\partial \hat m_t}$.
Note that LMP provides a new perspective for validating whether more transferable tickets can be discovered in pretrained models via tuning the sparsity patterns instead of model weights. 
Similar to that of the OMP case, robust tickets and natural tickets differ in whether they are drawn from adversarially or naturally trained dense models.

%% file: Sections/3-Experiment.tex
\begin{figure*}[t]
    \centering
    \includegraphics[width=\textwidth]{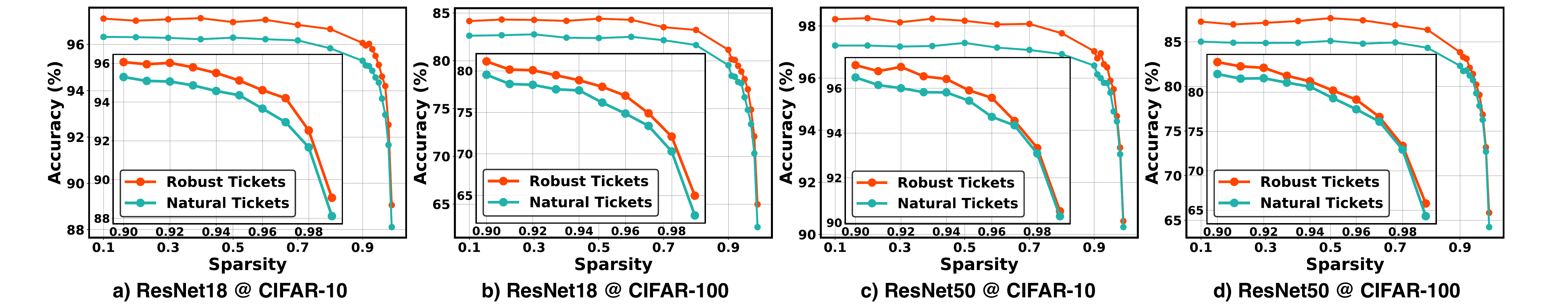}
        \vspace{-1.5em}
    \caption{Comparing the whole model finetuning accuracy of robust tickets and natural tickets identified via OMP from ResNet18/50 on CIFAR-10/100, with zoom-ins for the extreme sparsity ($90\%\sim99\%$).}
    \label{fig:omp-finetune}
    \vspace{-1.5em}
\end{figure*}

\begin{figure}[t]
    \centering
    \includegraphics[width=0.48\textwidth]{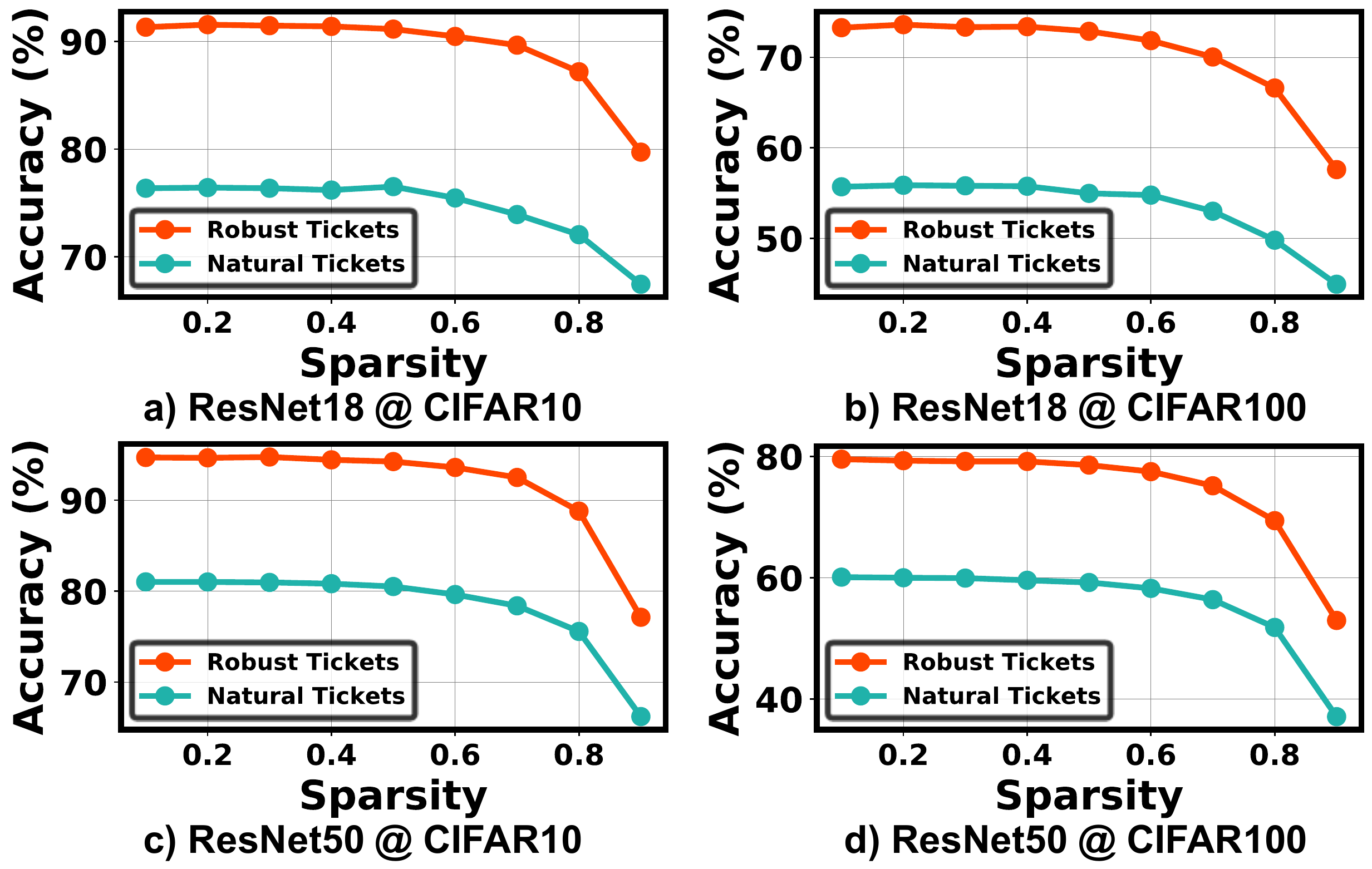}
        \vspace{-0.5em}
    \caption{Comparing the linear evaluation accuracy of robust tickets and natural tickets identified via OMP.}
    \label{fig:omp-linear}
      \vspace{-0.5em}
\end{figure}

\begin{figure}[h]
    \centering
    \includegraphics[width=0.48\textwidth]{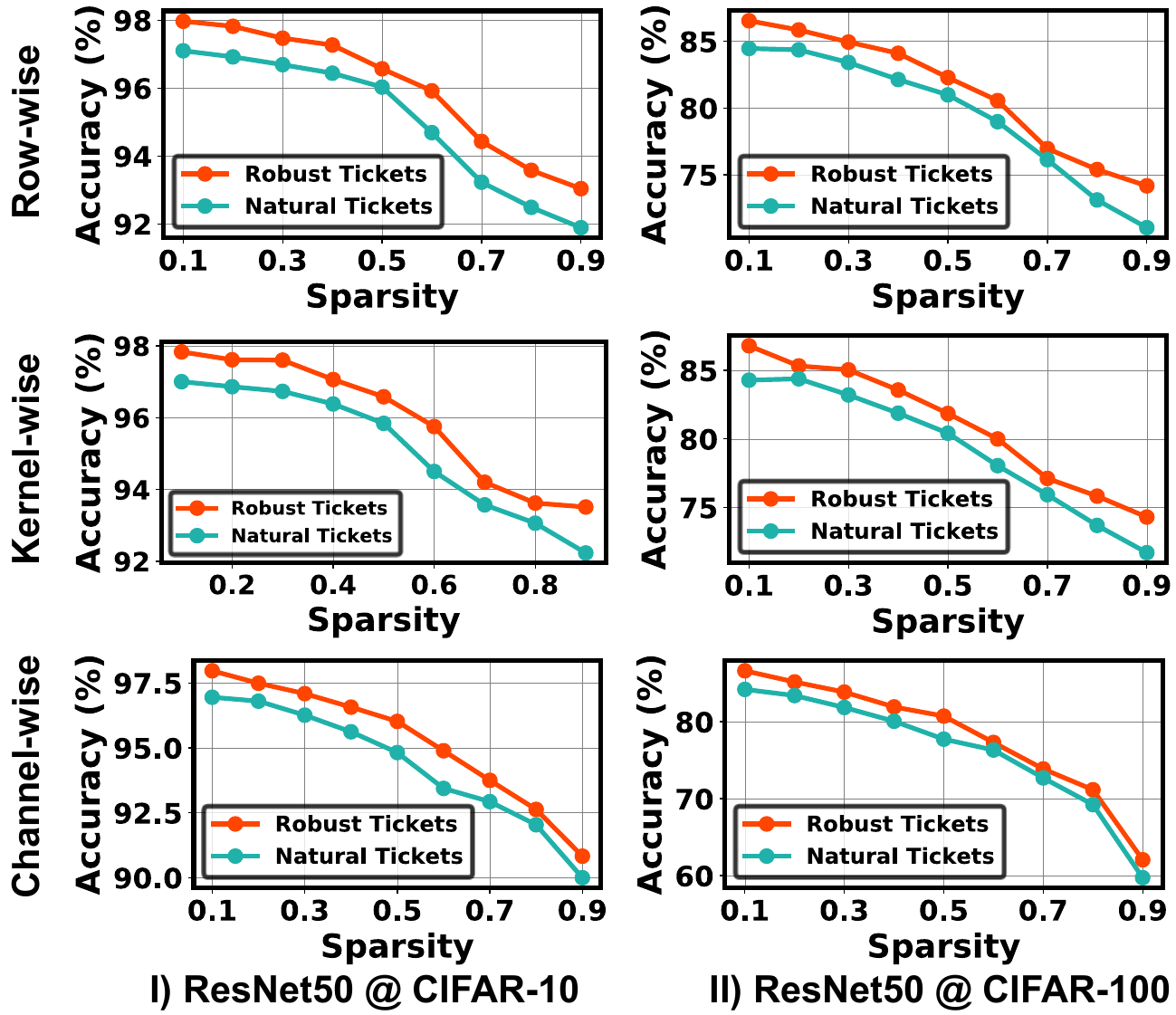}
        \vspace{-0.5em}
    \caption{Evaluating structured robust tickets over natural ones discovered via OMP from ResNet50.}
    \label{fig:omp-structured-r50}
    \vspace{-1.5em}
\end{figure}

\section{Experimental Results}
\label{sec:evaluation}

In this subsection, we aim to answer the research questions that are critical in better understanding and guiding practical uses of our proposed robust tickets via extensive experiments.

\subsection{Experiment Setup}
\textbf{Models and datasets.} We consider ResNet18 and ResNet50, which are commonly adopted feature extractors for transfer learning~\cite{chen2021lottery,salman2020adversarially}, featured by different degrees of overparameterization. We consider 14 datasets across different application domains, including classification on 13 datasets (CIFAR-10/100 and another 11 tasks from VTAB~\cite{zhai2019large}) and segmentation on PASCAL VOC~\cite{Everingham10}.

\textbf{Pretraining settings.} By default, we adopt ImageNet~\cite{russakovsky2015imagenet} as the source task for pretraining the models.
For robust pretraining, we adopt PGD training~\cite{madry2017towards} by default and we follow~\cite{salman2020adversarially} to pick the optimal perturbation strength for each task.

\textbf{Finetuning settings.}
For classification tasks, we follow the settings in~\cite{salman2020adversarially}, i.e., using an SGD optimizer for finetuning 150 epochs in total with a batch size of 64, a momentum of 0.9, and a weight decay of 1e-4. The learning rate decays by 0.1 at the 50-th and 100-th epochs. For segmentation on PASCAL VOC, we finetune for 30k iterations using an SGD optimizer with a momentum of 0.9, a weight decay of 1e-4, and a batch size of 4. The learning rate decays by 0.1 at the 18k-th and 22k-th iterations.

\begin{figure*}[t]
    \vspace{-1em}
    \centering
    \includegraphics[width=0.85\textwidth]{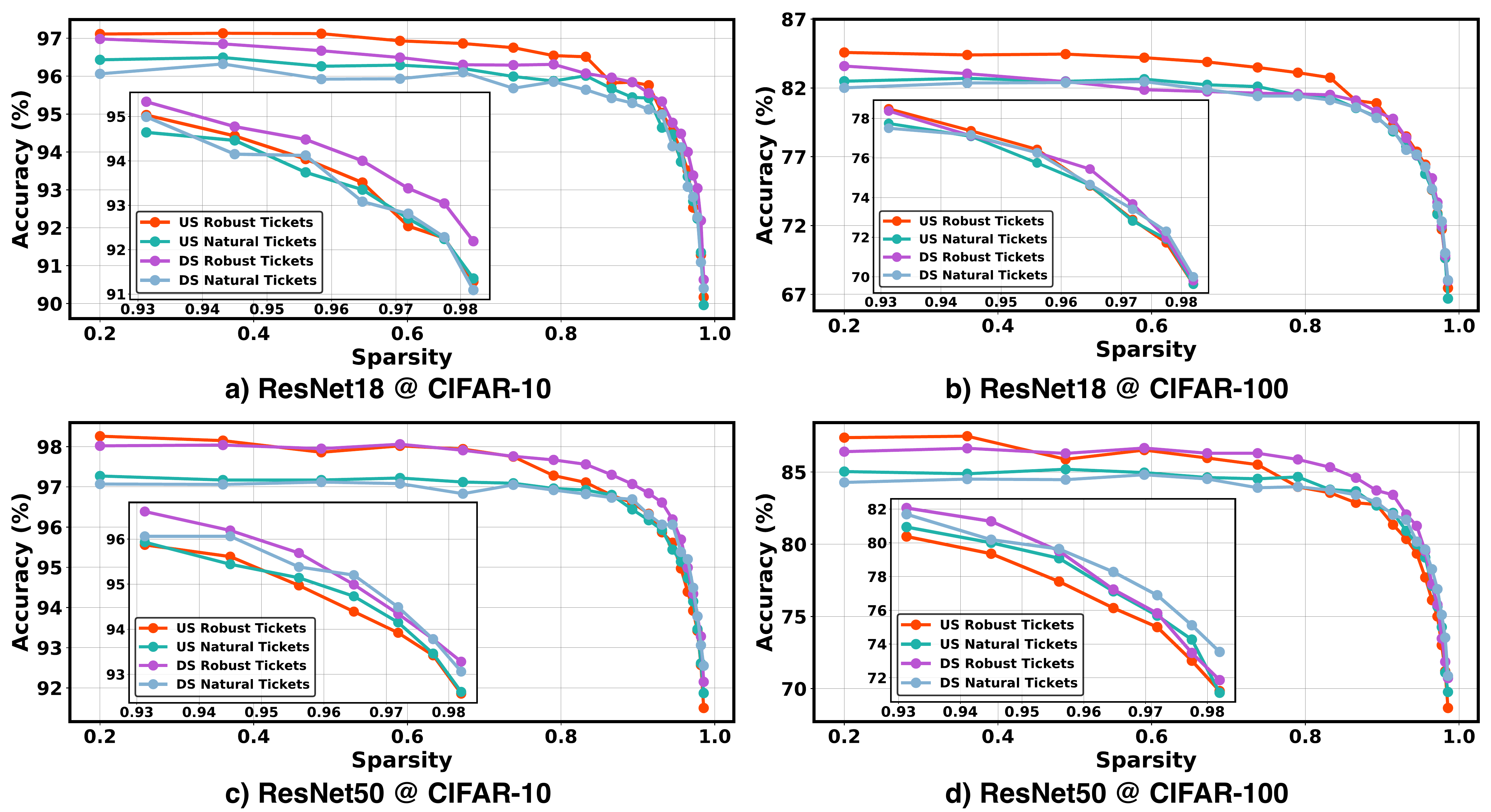}
    \vspace{-0.5em}
    \caption{Benchmark robust tickets with natural ones discovered by IMP variants. Accuracy under high sparsity is zoomed in.}
    \label{fig:Imp}
    \vspace{-1.5em}
\end{figure*}

\subsection{Whether robust tickets can transfer better across pruning methods and sparsity patterns?}

\textbf{Drawing robust tickets via OMP.}
We benchmark the transferability of robust tickets and natural tickets drawn by applying OMP on robust/naturally pretrained models. respectively, across various downstream classification tasks. We consider both whole-model finetuning and linear evaluation, where the weights of the drawn tickets are fixed and only a new classifier is learned on top of their extracted features.

\underline{Benchmark under whole-model finetuning.} We discover robust and natural tickets at different sparsity ratios from ResNet18/50, which are transferred to CIFAR-10/100 with whole-model finetuning. As shown in Fig.~\ref{fig:omp-finetune}, we can see that \underline{(1)} aligning with our hypothesis, robust tickets can consistently outperform the nature tickets under the whole-model finetuning paradigm, e.g., robust tickets achieve a 1.95\% higher accuracy on  ResNet50/CIFAR-100 under a sparsity of 0.7, and \underline{(2)} the robustness priors can be effectively inherited by extremely sparse subnetworks according to the consistent benefit of robust tickets over natural ones, e.g.,  a 2.38\% higher accuracy on  ResNet18/CIFAR-100 under a sparsity as high as 0.99.

\underline{Benchmark under linear evaluation.} As shown in Fig.~\ref{fig:omp-linear}, we can observe that robust tickets aggressively win the transferred accuracy, e.g., a  $\geqslant$11.75\% higher accuracy on ResNet50/CIFAR-100 till the sparsity ratio of 0.92. This indicates the superiority of robust tickets as a fixed feature extractor with notably better tolerance to potential domain shifts, thanks to the robustness priors.

\underline{Structured robust tickets.} We discover structured robust tickets, which benefit the real-hardware acceleration, via pruning the pretrained models at different granularities, including row-wise, kernel-wise, and channel-wise ones. As shown in Fig.~\ref{fig:omp-structured-r50}, we can see that \underline{(1)} robust tickets consistently win across different sparsity patterns and evaluation paradigms, and \underline{(2)} it is harder for more structured tickets to inherit robustness priors according to the smaller  gains over natural ones with more coarse-grained sparsity patterns.

\underline{Key insights.} This set of experiments indicates that \ding{182} robust tickets identified by the simplest OMP mechanism can already outperform the natural counterparts, which implies the general existence of highly transferable subnetworks in robust pretrained models, and \ding{183} although adversarial robustness can also serve as good priors for extremely sparse subnetworks, which benefits more efficient transfer learning, their advantages over natural ones are smaller than the ones under relatively low sparsity since fewer neurons during pretraining are inherited in the former case.

\textbf{Drawing robust tickets via A-IMP.}
We benchmark robust tickets derived by A-IMP with the natural ticket counterparts derived by vanilla IMP in Fig.~\ref{fig:Imp}, where `US' and `DS' denote whether A-IMP/IMP is performed on the upstream (source) task or the downstream task, respectively. Note that we derive robust and natural tickets from adversarially and naturally pretrained models, respectively. We can observe that \underline{(1)} generally the two robust tickets outperform the natural ones across most sparsity ratios; \underline{(2)} the US robust tickets achieves the most competitive transferability under mild sparsity and the DS robust tickets achieve a better accuracy over the US ones under relatively large sparsity, where updating the sparsity patterns via IMP based on task-specific information of the downstream task becomes more crucial; and \underline{(3)} for ResNet50 on CIFAR-100, the two robust tickets outperform natural ones under mild sparsity, whereas the latter becomes the winner under extremely high sparsity ($>$ 0.95) according to the zoom-in part of Fig.~\ref{fig:Imp} (d). We conjecture that this is because on more complex datasets, learning task-specific sparsity patterns via IMP becomes increasingly crucial under extremely high sparsity, where fewer robustness priors induced by pretraining can be inherited, favoring the natural tickets that benefit from more accurate pretrained models on the source task.

\begin{figure}[bht]
    \centering
    \includegraphics[width=0.48\textwidth]{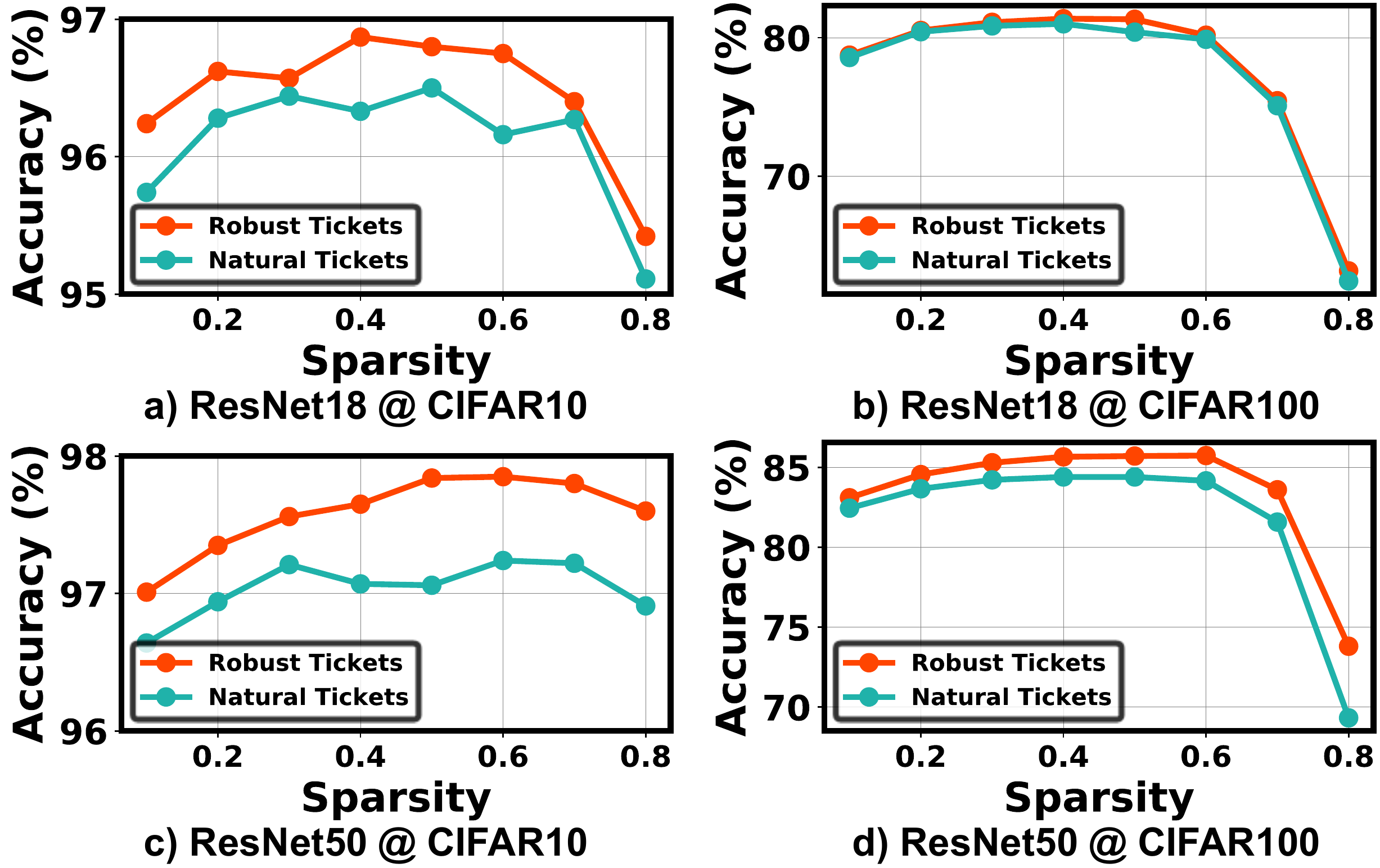}
    \caption{Benchmark robust tickets with natural tickets discovered by LMP from ResNet18/50 on CIFAR-10/100.}
    \label{fig:lmp}
    \vspace{-1.5em}
\end{figure}

\textbf{Drawing robust tickets via LMP.}
Different from OMP and A-IMP, the unique property of LMP is that only the learnable masks are optimized on downstream tasks while the model weights are fixed to the pretrained ones, thus LMP can be comprehended as directly extracting task-specific subnetworks hidden in the pretrained model. As shown in Fig.~\ref{fig:lmp}, robust tickets drawn by LMP consistently outperform natural ones, indicating that robust pretrained models with induced robustness priors are more likely to contain highly transferable subnetworks without weight finetuning, which validates our hypothesis from a new perspective.

\underline{Summary.} 
Comparing the aforementioned three pruning schemes for drawing robust tickets, IMP-based US/DS robust tickets generally win the best transferability, especially under extremely high sparsity ratios, which may be attributed to the robust training objective during the iterative optimization for discovering the robust tickets.

\begin{figure}[t]
    \centering
    \includegraphics[width=0.48\textwidth]{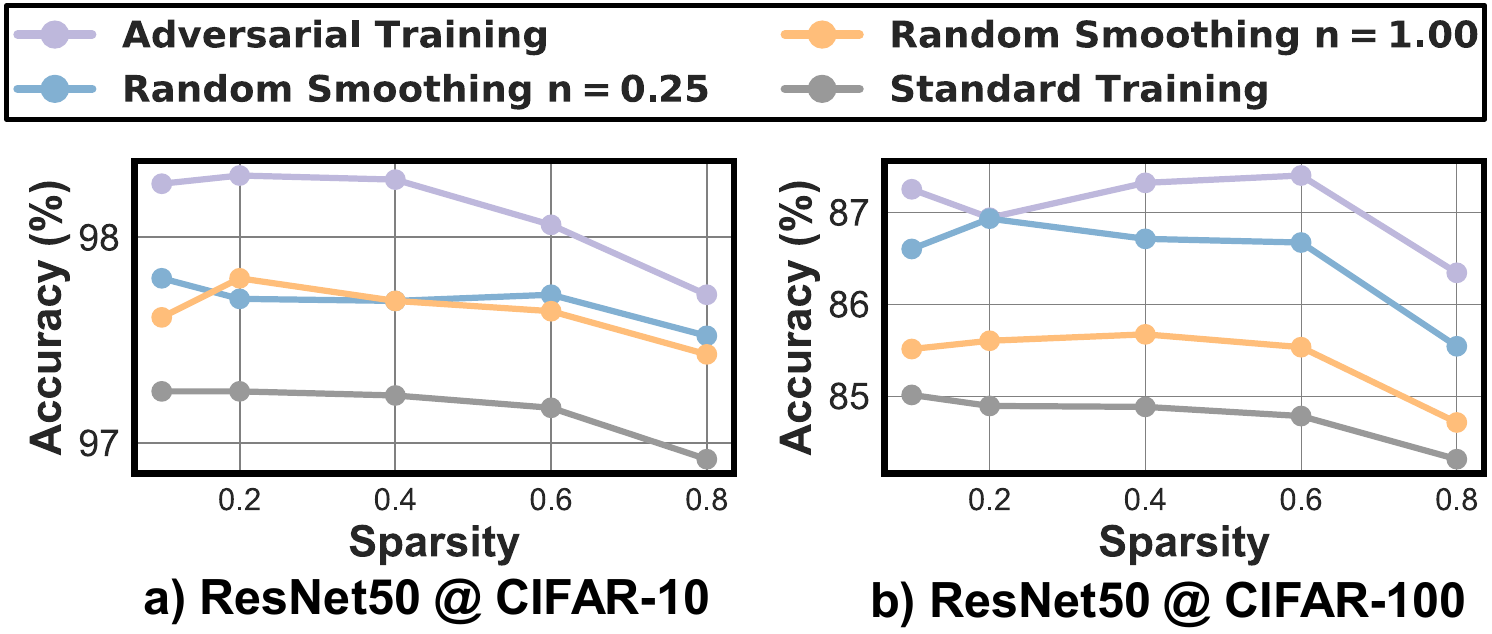}
    \caption{Benchmark the tickets from different pretraining schemes.}
    \label{fig:pretrain_scheme}
    \vspace{-2em}
\end{figure}

\vspace{-0.5em}
\subsection{Whether adversarial pretraining is necessary for robust tickets?}
\label{sec:adv_pretraining}
\textbf{Setup.}
To study whether adversarial training is the only way to induce robustness priors for enabling better transferability, we apply random smoothing (RS)~\cite{cohen2019certified} as an alternative robust training method for pretraining a ResNet50, on top of which the OMP is applied. 

\textbf{Results and analysis.}  We benchmark the transferability of different tickets drawn by OMP from naturally pretrained models and adversarially/RS pretrained models. As shown in Fig.~\ref{fig:pretrain_scheme}, we can observe that although the tickets from RS pretrained models are inferior to the robust tickets from adversarially trained models, they can still outperform the natural tickets.

\underline{Key insights.} This set of experiments indicates that robustness priors induced into pretraining via various robust training algorithms can be generally inherited by the drawn tickets and thus benefit the transferability to downstream tasks.

\begin{figure}[h]
    \vspace{-0.5em}
    \centering
    \includegraphics[width=0.45\textwidth]{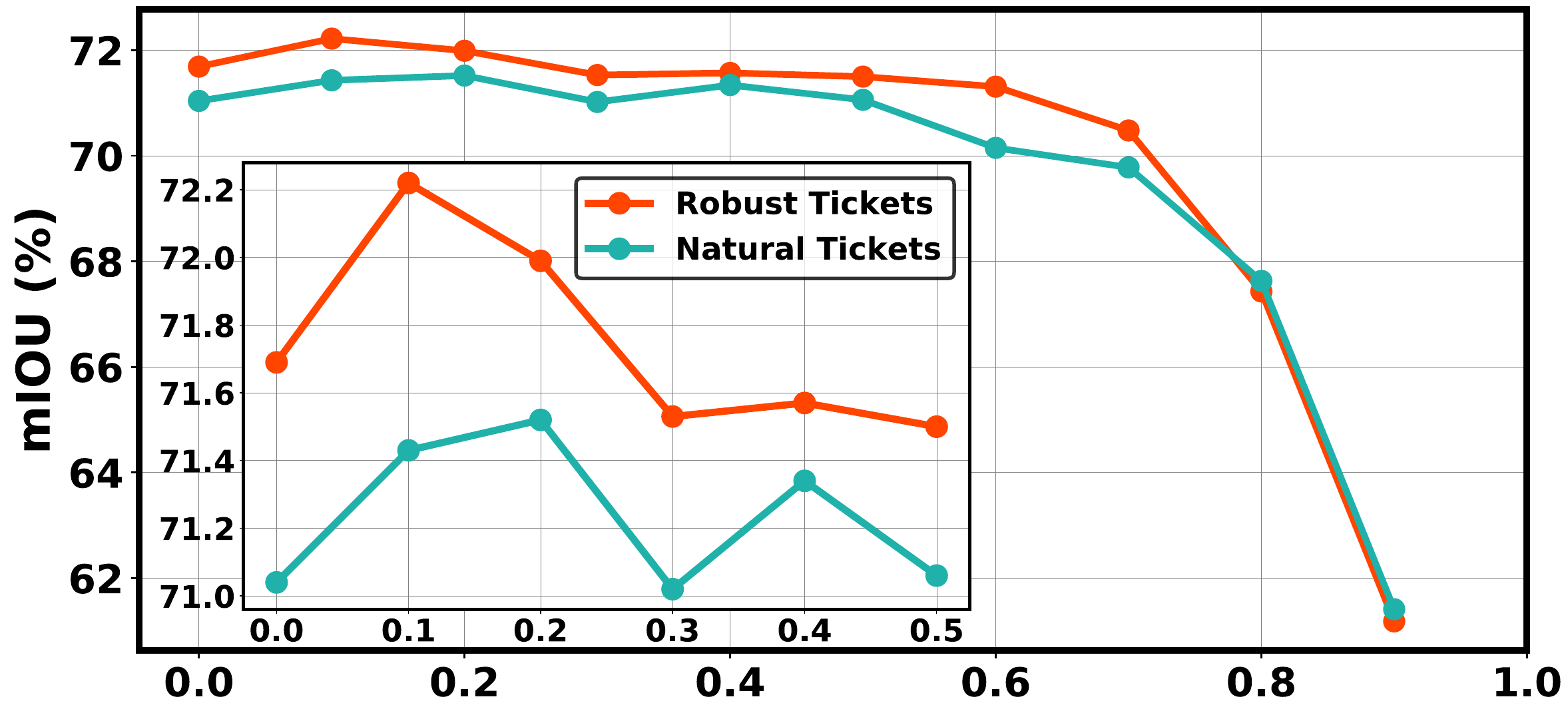}
        \vspace{-0.5em}
    \caption{Benchmark robust tickets and natural ones drawn from ResNet50 via OMP on the segmentation task.}
    \label{fig:voc}
    \vspace{-1.5em}
\end{figure}

\begin{figure*}[ht]                                  
    \vspace{-0.5em}
    \centering
    \includegraphics[width=0.92\textwidth]{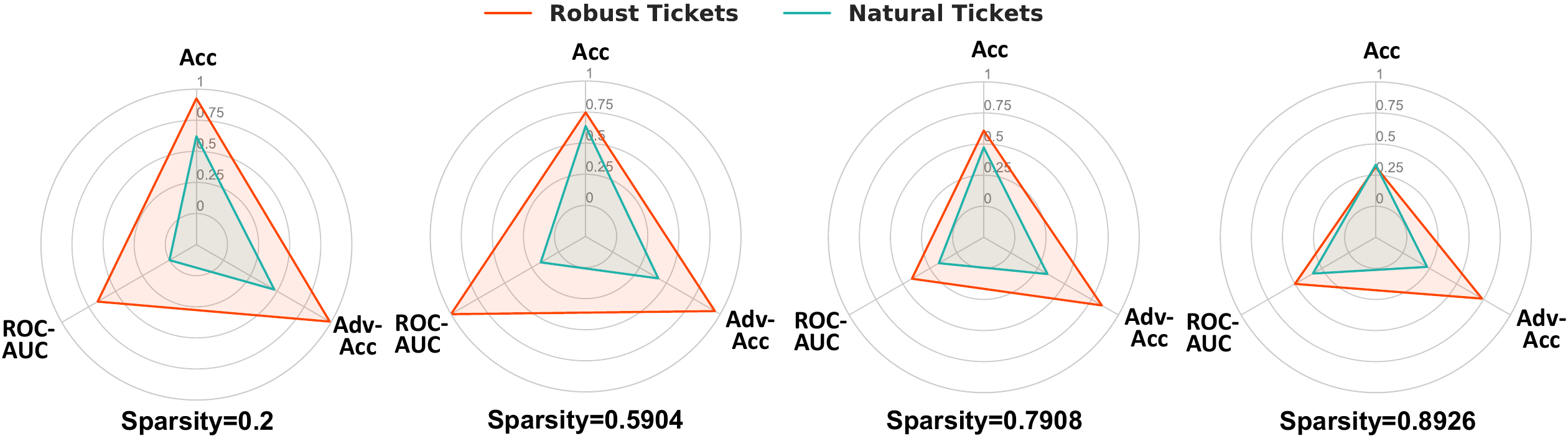}
    \caption{Benchmark robust tickets drawn by A-IMP and natural tickets drawn by IMP in terms of other properties, i.e. natural accuracy (denoted as Acc), robustness to adversarial perturbation (denoted as Adv-Acc) and OoD performance (denoted as ROC-AUC).} 
    \label{fig:imp-other}
    \vspace{-0.5em}
\end{figure*}

\begin{figure*}[t]
    \centering
    \includegraphics[width=0.92\textwidth]{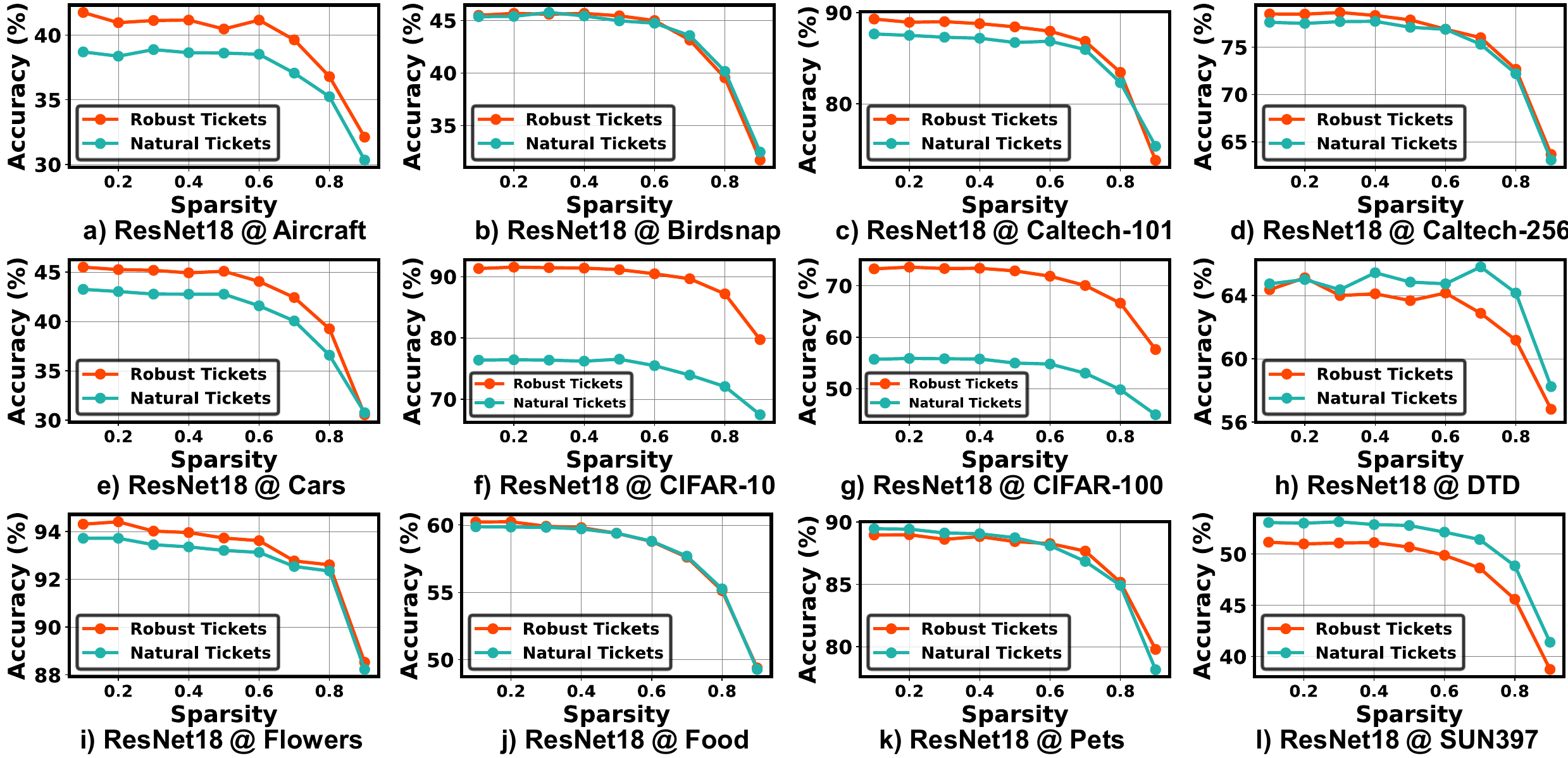}
    \caption{Comparing the linear evaluation accuracy of robust tickets and natural tickets identified via OMP method on 12 tasks from the VTAB benchmark~\cite{zhai2019large}. All tickets in this figure are identified from dense ResNet18 models.}
    \label{fig:omp-linear-VTAB-r18}
    \vspace{-1em}
\end{figure*}

\subsection{Will the conclusions be scalable to other downstream tasks?}

\textbf{Setup.} We transfer the OMP pruned robust tickets and natural tickets to the segmentation task on PASCAL VOC as shown in Fig.~\ref{fig:voc}.

\textbf{Results and analysis.} We can observe that generally our hypothesis still holds that robust tickets achieve consistently higher mIOU, especially under mild sparsity ratios, which indicates that the robustness priors could enhance the transferability across different downstream tasks, not limited to classification tasks. 

\underline{Key insights.} As implementing the adversarial pretraining on large-scale datasets is costly, we expect that our robustness priors could universally transfer to different downstream tasks. This set of experiments indicates that although the cost of adversarial pretraining is higher, the universal transferability of robustness priors across tasks could amortize this additional cost.

\subsection{What about other properties of robust tickets?} \textbf{Setup.} In addition to the natural accuracy, we also illustrate the out-performance of robust tickets over natural ones under different metrics, including adversarial robustness and OoD performance.

\textbf{Results and analysis.} From Fig.~\ref{fig:imp-other}, we can see that \underline{(1)} robust tickets outperform the vanilla counterparts consistently in terms of natural accuracy, adversarial accuracy, and OoD performance. This indicates that \ding{182} robustness priors can boost the capability to combat input perturbations in addition to enhancing the transferability, and \ding{183} robustness priors can improve large models' (e.g., ResNet50) OoD performance and thus enhance the reliability. The raw data for visualizing Fig.~\ref{fig:imp-other} is provided in Tab.~\ref{tab:raw}.

\begin{table}[ht]
\centering
\vspace{0.5em}
\caption{Raw data of all the mentioned properties of robust tickets and natural tickets drawn from ResNet18/50 via A-IMP.}
\resizebox{0.48\textwidth}{!}{
\begin{tabular}{cc|cccc|cccc}
\toprule
\multirow{2}[2]{*}{\textbf{Model}} & \multirow{2}[2]{*}{\textbf{Properties}} & \multicolumn{4}{c|}{\textbf{Robust Tickets}} & \multicolumn{4}{c}{\textbf{Natural Tickets}} \\ \cmidrule{3-10}
 &  & 20.00\% & 59.04\% & 79.08\% & 89.26\% & 20.00\% & 59.04\% & 79.08\% & 89.26\% \\ \midrule
\multirow{5}{*}{R-18} & Accuracy ↑ & 96.3 & 96.34 & 96.13 & 95.31 & 95.79 & 95.47 & 95 & 94.74 \\
 & ECE ↓ & 0.0159 & 0.015 & 0.0186 & 0.0201 & 0.0125 & 0.0173 & 0.0175 & 0.0264 \\
 & NLL ↓ & 0.1737 & 0.1769 & 0.1872 & 0.2166 & 0.1866 & 0.1954 & 0.2116 & 0.2336 \\
 & Adv-Acc ↑ & 55.83 & 53.9 & 51.93 & 47.69 & 27.89 & 25.71 & 23.43 & 19.9 \\
 & ROC-AUC ↑ & 0.83 & 0.74 & 0.73 & 0.66 & 0.89 & 0.83 & 0.85 & 0.8 \\ \midrule 
\multirow{5}{*}{R-50} & Accuracy ↑ & 97.6 & 97.05 & 96.62 & 95.72 & 96.66 & 96.71 & 96.2 & 95.77 \\
 & ECE ↓ & 0.0137 & 0.0132 & 0.0161 & 0.0179 & 0.0095 & 0.0121 & 0.0127 & 0.0128 \\
 & NLL ↓ & 0.1279 & 0.1457 & 0.1728 & 0.2028 & 0.1451 & 0.1509 & 0.1659 & 0.1961 \\
 & Adv-Acc ↑ & 67.4 & 65.62 & 60.59 & 55.32 & 42.66 & 40.34 & 36.3 & 30.79 \\
 & ROC-AUC ↑ & 0.8 & 0.84 & 0.77 & 0.78 & 0.72 & 0.74 & 0.74 & 0.76 \\ \bottomrule
\end{tabular}
}
 \label{tab:raw}
 \vspace{-1.5em}
\end{table}

\subsection{When and why could robust tickets transfer better?}

\textbf{Setup.} To understand the underlying reasons behind robust tickets' transferability as well as identify when robust tickets win, we extend the linear evaluation setting to 12 tasks in VTAB~\cite{zhai2019large}. Here we also measure the FID score~\cite{heusel2017gans}, which indicates the difference in data distribution between two datasets (note that a lower FID denotes a smaller domain gap), between the source dataset ImageNet and each downstream dataset in VTAB by sampling the same 8000 images from ImageNet and using all the images in each downstream dataset.

\textbf{Observations and analysis.} 
As shown in Fig.~\ref{fig:omp-linear-VTAB-r18}, robust tickets outperform/match/underperform natural tickets in 7/3/2 out of the total 12 cases under high sparsity ratios. According to Tab.~\ref{tab:fid}, we can observe that robust tickets consistently outperform natural ones on datasets with a larger FID, where the domain gap between downstream datasets and ImageNet is larger, and only match/underperform their natural counterparts on datasets with a lower FID. This indicates that adversarial robustness can serve as a prior for transferability as it reflects the capability for dealing with domain gaps and thus could be helpful in real-world applications where the distributions of target tasks often diverge from those of the source tasks.

\begin{table}[h]
  \centering
  \caption{Winning tickets on each dataset in VTAB with the corresponding FID calculated over ImageNet.}
\vspace{-0.5em}
  \resizebox{0.48\textwidth}{!}{
    \begin{tabular}{ccccccc}
    \toprule
    \textbf{Dataset} & \textbf{CIFAR-10} & \textbf{Aircraft} & \textbf{CIFAR-100} & \textbf{Pets}   & \textbf{Flowers}     & \textbf{Cars}        \\ \midrule
    \textbf{FID Score} & 205.04 & 198.33 & 190.31 & 173.23 & 153.76      & 150.92 \\ 
    \textbf{Winner}  & \textbf{Robust}       & \textbf{Robust}       & \textbf{Robust}        & \textbf{Robust}     & \textbf{Robust}          & \textbf{Robust}          \\ \midrule \midrule
    \textbf{Dataset} & \textbf{Food}     & \textbf{DTD}      & \textbf{Birdsnap}  & \textbf{SUN397} & \textbf{Caltech-101} & \textbf{Caltech-256} \\ \midrule
    \textbf{FID Score} & 115.95 & 97.33  & 92.64  & 67.7   & 56.71       & 27.54  \\
    \textbf{Winner}    & Match  & Natural     & Match  & Natural     & \textbf{Robust} & Match  \\ \bottomrule
    \end{tabular}
  }
  \vspace{-0.5em}
  \label{tab:fid}%
\end{table}%

\underline{Key insights.}
This experiment complements our key insights and enhances our understanding of when robust tickets perform better, which also indicates that FID is one potential metric to choose the proper transfer learning scheme.

%% file: Sections/4-Related-Work.tex
\section{Related Work}
\label{sec:related_work}

\textbf{Lottery ticket hypothesis.}
The lottery ticket hypothesis~\cite{frankle2018lottery} is the first to discover that there exist trainable subnetworks which are hidden in an over-parameterized dense network but can match the accuracy of their dense network counterparts after being trained in isolation. This intriguing finding has motivated various follow-up works that explore the lottery ticket hypothesis under different scenarios. 
In particular,~\cite{chen2021lottery,iofinova2021well} extend the lottery ticket hypothesis to transfer learning and find that subnetworks inheriting the pretrained model weights as initialization can match the task accuracy of their dense network counterparts after finetuning. However, their adopted metrics were originally designed for maintaining the accuracy of the identified tickets on the same task and thus do not necessarily help preserve the transferability to downstream tasks.

\textbf{Transfer learning.}
Motivated by the success of DNNs as a general feature extractor, transfer learning \cite{sharif2014cnn} has been widely adopted across different domains to benefit  downstream tasks from DNNs pretrained on a big data regime. 
Although pioneering
works~\cite{kornblith2019better} have investigated various factors that can affect the effectiveness of transfer learning, how to discover highly transferable and sparse subnetworks is still an open research question.

\textbf{Adversarial robustness.}
DNNs are known to be vulnerable to adversarial attacks~\cite{goodfellow2014explaining}.
As adversaries, stronger attacks have been continuously proposed to degrade the accuracy of target DNNs, including both white-box~\cite{madry2017towards} and black-box ones~\cite{andriushchenko2020square}. In response, various defense methods have been proposed to enhance DNNs' adversarial robustness~\cite{shafahi2019adversarial, madry2017towards}. 
In addition, adversarial robustness has been found to be highly correlated with other desired properties, e.g., reducing overfitting for image recognition~\cite{xie2020adversarial} and enabling better transfer learning~\cite{salman2020adversarially,deng2021adversarial}. Recent works~\cite{fu2021double, fu20212} also aim to achieve both robustness and efficiency within a single framework.

%% file: Sections/5-Conclusion.tex
\section{Conclusion}
\label{sec:conclusion}

Recent advances in transfer learning have provided a promising data-efficient solution for enhancing the achievable task performance of downstream tasks. 
To further enhance the model efficiency towards wide-scale adoption, we study what good priors are for identifying highly transferable subnetworks on top of the lottery ticket hypothesis and interestingly discover that adversarial robustness can serve as a good prior for drawing more transferable subnetworks. We conduct extensive experiments across diverse tasks, sparsity patterns, pretraining schemes, and performance metrics to understand the properties of robust tickets and analyze the underlying reasons behind their transferability. 
Our work has complemented the lottery ticket hypothesis and opened up a new perspective for empowering transfer learning on edge devices by pushing forward the achievable transferability-sparsity trade-offs.

\section*{Acknowledgement}
The work is supported by the National Science Foundation (NSF) through the MLWiNS program (Award number: 2003137) and the EPCN program (Award number: 1934767).